\documentclass[12pt]{article}

\usepackage[margin=1in]{geometry}
\usepackage{setspace}
\usepackage{booktabs}
\usepackage{longtable}
\usepackage{natbib}
\usepackage[hidelinks]{hyperref}
\usepackage{indentfirst}
\usepackage{pdflscape}
\usepackage{array}
\usepackage{caption}
\usepackage{graphicx}
\usepackage{subcaption}
\title{Improving the Completeness and Comparability of Segment Disclosures: A Large Language Model Approach}

\author{
{\small Yue Liu} \\
{\small Rutgers Business School} \\
{\small Rutgers University - Newark} \\
{\small Newark, NJ, USA} \\
{\small \texttt{yl1641@scarletmail.rutgers.edu}}
\and
{\small Zhiyuan Cheng} \\
{\small School of Engineering} \\
{\small Stanford University} \\
{\small Stanford, CA, USA} \\
{\small \texttt{zhycheng@stanford.edu}}
\and
{\small Longying Lai} \\
{\small Simon Business School} \\
{\small University of Rochester} \\
{\small Rochester, NY, USA} \\
{\small \texttt{longyinglai703@gmail.com}}
}

\date{\small Apr 16, 2026}

\begin{document}

\maketitle

\newpage
\begin{abstract}
Segment-level disclosures are a central component of financial reporting, providing insight into firms’ internal organization and the allocation of economic activities across operating units. However, segment information is often presented in both qualitative and quantitative forms, dispersed across tables and narrative sections of Form 10-K filings. Empirical research relying on structured databases faces both completeness and comparability challenges, as some firm-year observations may be missing, nested segment disclosures are not captured, and support for longitudinal and cross-firm comparability is limited. This study develops a large language model–based framework to extract segment disclosures directly from Form 10-K filings and to preserve both reportable and nested segment information. We further design a retrieval augmented system that incorporates information across multiple filings to support comparability. We use two representative settings to demonstrate its application: longitudinal analysis within a firm to interpret segment changes over time, and cross firm alignment of geographic segments across firms with different reporting structures. The results indicate that the artifact accurately extracts segment-level information and effectively addresses questions that require cross-period knowledge, demonstrating the potential of LLM-based approaches to enhance the measurement and interpretation of segment disclosures.
\end{abstract}

\noindent \textbf{Data Availability: } Data sets are available from the public sources cited in the text. 
\noindent \textbf{Financial Conflicts: } The authors declare no financial conflicts of interest.

\newpage
\section{Introduction}

Segment disclosure is an important component of corporate financial reporting, providing investors and other stakeholders with insight into how firms’ operations, performance, and risks vary across business activities. By disaggregating consolidated financial information, segment disclosures allow users to better understand the sources of firm value and to evaluate managerial decision-making through the lens of internal organization. As a result, segment-level information has long been a focus of accounting regulation and empirical research.

Segment disclosure requirements under U.S. GAAP have evolved from rule-based classification toward a management-oriented reporting framework. Early guidance was established by SFAS No. 14\citep{fasb1976}, which required firms to report industry and geographic segments using externally defined criteria\footnote{External criteria refer to standardized classification schemes defined outside the firm, such as industry classifications based on Standard Industrial Classification or similar externally imposed guidelines. Under SFAS No. 14, segment identification was guided by these predefined categories rather than firms’ internal organizational structure, resulting in more uniform but less management relevant disclosures.}. However, evidence that SFAS 14 allowed excessive aggregation and produced disclosures with limited decision usefulness \citep{street2000} led the Financial Accounting Standards Board to replace it with SFAS No. 131 \citep{fasb1997}. SFAS 131 introduced the management approach, requiring firms to disclose operating segments based on internal organizational structures and information regularly reviewed by the chief operating decision maker. This approach was intended to enhance the relevance of segment information by aligning external reporting with internal performance evaluation and resource allocation decisions. SFAS 131 was subsequently codified in ASC Topic 280\citep{fasb2023a,fasb2023b}, which continues to govern segment reporting in the United States. Under ASC 280, firms must disclose segment revenues, profit or loss measures, assets, reconciliations to consolidated totals, and selected geographic and major customer information when applicable. While the management approach improves informational relevance, it also grants managers substantial discretion in defining segment boundaries, aggregation levels, and descriptive labels, contributing to heterogeneity and comparability challenges in segment disclosure \citep{berger2003, ,hinson2019a,hinson2019b}.

\begin{figure}[p]
\centering
\caption{AAPL Segment Disclosure for FY2024}
\label{fig:fig1}
\includegraphics[width=\textwidth]{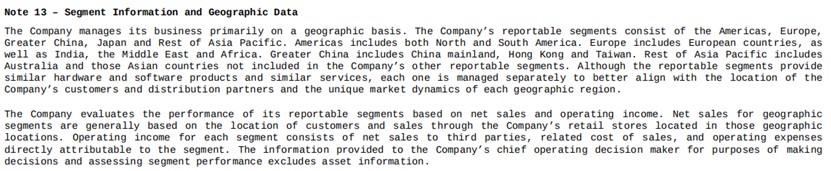}
\vspace{-0.15cm}
\includegraphics[width=\textwidth]{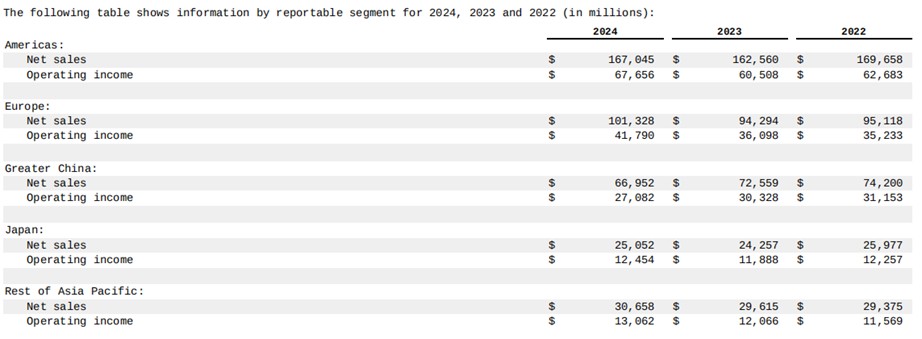}
\end{figure}

These features create challenges for large sample studies. First, although segment disclosures include tabular presentations, important information is often distributed across both tables and narrative footnotes within Form 10-K filings, as illustrated in Figure~\ref{fig:fig1}. This dispersion makes it difficult to systematically extract and integrate segment information at scale. Second, the managerial discretion introduced under the management approach leads to variation in how segments are defined, aggregated, and labeled, creating inconsistencies both across firms and within the same firm over time. Such variation complicates the identification and comparison of segment structures. As a result, commonly used structured databases may contain missing observations, omit relevant details, or introduce inconsistencies that obscure the underlying economic content of segment disclosures.

Recent advances in large language models (LLMs) offer new opportunities to address these challenges. LLMs have demonstrated strong capabilities in processing long, unstructured textual documents and extracting structured information from heterogeneous sources \citep{zhang2025}. Unlike rule based or traditional natural language processing methods, LLMs can capture contextual relationships across sentences, link information presented in different parts of a document, and interpret variations in wording and presentation. These capabilities are particularly well suited to segment disclosures, which require integrating information from both narrative explanations and footnote tables and accommodating substantial variation in disclosure style across firms and years. In addition, recent evidence suggests that LLMs can extend beyond textual understanding to support quantitative financial tasks, including multi-step reasoning and strategy generation \citep{kang2026}. This capability is particularly important in our setting, as comparing segment disclosures across firms with different reporting structures often requires reconciling definitions and performing basic aggregation and transformation of financial quantities.

This paper develops a large language model based framework to extract and interpret segment disclosures directly from firms’ Form 10 K filings. The approach is designed to address two fundamental challenges in segment level research: incompleteness and comparability. To address incompleteness, we construct a file grounded extraction procedure that recovers both reportable segments and nested disclosures from raw filings, preserving the hierarchical structure of segment reporting that is often lost in structured databases. By working directly with source documents, the framework captures information that is dispersed across narrative explanations and tabular presentations, providing a more complete representation of firms’ segment disclosures.

To improve comparability in firm segment level disclosure, we extend the framework with a retrieval augmented component that incorporates information across multiple filings. This design allows segment disclosures to be interpreted in context rather than in isolation. We consider two representative settings. First, in longitudinal analysis within a firm, the approach enables the identification of segment changes, summarizes the economic reasons underlying these changes, and links the current segment structure to prior years. Second, in cross firm analysis, we compare geographic segment disclosures by examining revenue in Asia for two firms within the same industry that use different aggregation levels. Instead of relying on predefined rules or structured identifiers in databases such as Compustat Historical Segment, the approach uses contextual information in disclosures to reconcile differences in naming, aggregation, and reporting structure. Together, these components provide a unified framework for constructing segment data that is both more complete and more comparable, facilitating more reliable use of segment information in archival accounting research.

The remainder of the paper proceeds as follows. Section 2 reviews the related literature. Section 3 identifies the measurement challenges in segment disclosures that motivate the design of the artifact. Section 4 describes the development and implementation of the LLM-based extraction and retrieval system. Section 5 evaluates the performance of the artifact and presents validation results. Section 6 concludes and discusses limitations and avenues for future research.

\section{Literature Review}

\subsection{Managerial Discretion in Segment Disclosure}

Segment reporting under the management approach grants managers substantial discretion in defining, organizing, and labeling operating segments. SFAS No. 131, later codified in ASC Topic 280, requires firms to report segment information based on internal reports reviewed by the chief operating decision maker (CODM), rather than imposing externally defined industry classifications. This design was intended to enhance informational relevance by aligning external segment disclosures with internal resource allocation and performance evaluation processes.

Prior research demonstrates, however, that the management approach also introduces meaningful discretion into segment disclosure. \citet{berger2003}show that the transition from SFAS No. 14 to SFAS No. 131 increased segment disaggregation and altered the informativeness of disclosures, consistent with managerial incentives influencing how segments are defined and reported. \citet{bens2011} provide evidence that managers exercise discretion in aggregating operating segments, particularly when disaggregation could reveal proprietary information or reduce reporting flexibility. Similarly, \citet{wang2015}documents that managers use discretion in allocating corporate-level expenses and income across segments, affecting reported segment profitability.

Beyond measurement choices, research also highlights cross-firm heterogeneity and comparability challenges arising from discretionary segment definitions. \citet{nichols2013} review the literature on the management approach and conclude that while segment disclosures may better reflect internal operations, they also reduce standardization across firms. Because operating segments are defined internally and may evolve with organizational restructuring, firms differ substantially in aggregation levels, naming conventions, and segment performance measures, even when underlying economic activities are similar.

An important implication of this discretion is that segment disclosures are not static reporting constructs. Firms may rename segments, modify descriptive labels, or re-aggregate business units without necessarily undergoing substantive changes in underlying operations. Such flexibility complicates longitudinal analysis and introduces ambiguity when researchers attempt to interpret segment additions, eliminations, or restructurings based solely on reported segment identifiers. This inherent discretion in segment reporting creates measurement challenges that motivate closer examination of primary disclosure documents.

\subsection{Measurement Challenges and Limitations of Structured Databases}

Because segment disclosures are typically presented within narrative footnotes and customized tabular formats in Form 10-K filings\footnote{\citet{cfainstitute2018} of financial statement users notes that U.S. GAAP does not prescribe a uniform presentation format for segment disclosures and that segment information is commonly embedded within narrative footnotes rather than presented in standardized tables. The report highlights that this flexibility contributes to variation in how firms present segment results and related reconciliations. Available from: https://rpc.cfainstitute.org/research/surveys/segment-disclosures-survey-report? }\footnote{\citet{deloitte2025} observes that segment disclosures, including enhancements under ASU 2023-07, are shaped by each registrant’s internal reporting practices, resulting in customized presentation formats across firms. Available from: https://dart.deloitte.com/USDART/home/publications/deloitte/financial-reporting-spotlight/2025/disclosure-trends-2024-reporting-season.}, empirical research frequently relies on structured commercial databases, most prominently Compustat, to obtain segment-level information. While these databases have facilitated a substantial body of research, prior accounting literature cautions that variables derived from secondary data sources may deviate from underlying disclosures due to data processing conventions, classification rules, aggregation procedures, and reporting heterogeneity.

\citet{hribar2002} demonstrate that commonly used accounting measures can be affected by mechanical data construction choices, leading to measurement error when researchers rely exclusively on database fields. Similarly, \citet{loughran2014} show that standardized textual processing methods may fail to capture context-specific disclosure characteristics embedded in narrative filings. These broader concerns are particularly salient for segment reporting, where disclosures are inherently firm-specific and shaped by managerial discretion under ASC Topic 280.

Segment-level databases therefore face distinct measurement challenges. First, coverage may be incomplete. For earlier periods, especially prior to the SEC’s XBRL mandate, segment disclosures were not machine-readable, requiring manual extraction or third-party standardization. Even in later periods, structured datasets may omit firm-year observations or nested disclosures embedded within segment footnotes. Second, segment identifiers and descriptive labels often vary across firms and over time. Because structured databases typically rely on mechanical string matching or internally assigned identifiers, minor naming differences, aggregation changes, or formatting variations can fragment longitudinal records and introduce noise into within-firm analyses.

Taken together, these limitations suggest that structured databases provide a processed representation of segment disclosures rather than a direct reflection of primary filings. When disclosures are discretionary, hierarchical, and textually embedded, reliance solely on standardized database fields may obscure economically meaningful variation. These concerns motivate approaches that return to original Form 10-K filings to reconstruct segment information directly from source documents.

\subsection{Extracting Information from Financial Disclosures}

Although Form 10-K filings follow a standardized reporting structure defined by the U.S. Securities and Exchange Commission (SEC), much of the information within these filings is presented in semi structured formats, including narrative text, customized tables, and footnotes that vary across firms and reporting periods. To facilitate automated retrieval of financial information, the SEC introduced a mandate requiring registrants to submit financial statement data using eXtensible Business Reporting Language (XBRL), which tags individual financial statement items with standardized taxonomy elements \citep{sec2009}. XBRL substantially improves the machine readability of financial statement data. However, empirical evidence suggests that tagging inconsistencies, taxonomy extensions, and mapping differences can lead to discrepancies between XBRL extracted values and the underlying disclosures in SEC filings\citep{chychyla2015}.

To address these limitations, \citet{zhang2023}apply natural language processing techniques to extract information from portions of Form 10-K filings that are not fully captured by structured tags. They develop a Form 10-K Itemization system that automatically partitions filings into standardized SEC Item sections and converts lengthy reports into structured components suitable for downstream information extraction tasks. By isolating semantically meaningful sections prior to analysis, such approaches enable systematic extraction of information embedded in narrative disclosures that are not available through XBRL tagged data.

Recent advances in large language models (LLMs) extend these extraction pipelines by providing context-aware, flexible information retrieval from heterogeneous and unstructured financial disclosures\citep{liu2024a, yao2025, lai2026}. LLMs can integrate narrative text, tables, and cross-sectional contextual cues, enabling more accurate extraction of structured variables from complex and variably formatted filings.\citet{li2025} develop an LLM-based extraction framework that successfully recovers financial data from unstructured reports with high documented accuracy, providing direct evidence that LLMs can overcome limitations of prior rule- or NLP-based extraction approaches. Such models are particularly promising for segment disclosure extraction, where disclosures combine narrative explanations with customized textual interpretation to reconstruct segment-level data.

\subsection{Retrieval Augmented Generation in Financial Question Answering}
Retrieval Augmented Generation (RAG) has emerged as a prominent framework for enhancing large language models in knowledge intensive tasks\citep{shi2026}. A central motivation for adopting RAG is the limitation of directly providing full documents to a large language model \citep{shi2025}. Although recent models support long context windows, prior studies show that model performance does not scale proportionally with context length, and relevant information may be underutilized when embedded in long inputs\citep{liu2024b,fang2025,hu2026}.The core idea of RAG is to combine a retrieval component with a generative model, allowing the system to access external documents at inference time rather than relying solely on parametric knowledge learned during training. In the original formulation, the retriever identifies relevant passages from a corpus and the generator produces responses conditioned on both the query and the retrieved content\citep{lewis2020}. Subsequent research has expanded this framework and demonstrated its effectiveness across a wide range of applications, particularly in domains that require accurate grounding in source documents\citep{gao2024}.

Answering financial questions is one of the scenarios in which Retrieval Augmented Generation can be particularly helpful, because relevant information is often distributed across multiple sections and multiple documents rather than contained within a single file. Financial disclosures such as Form 10 K filings are lengthy and complex, and many questions require combining information across reporting periods or across different parts of the filing. In such settings, directly providing all documents to a large language model is inefficient and may lead to information overload or incorrect associations. RAG addresses this issue by retrieving only the most relevant information and incorporating it into the generation process. The FinDER dataset, which is constructed from real world analyst queries over Form 10-K reports, provides a realistic setting to evaluate this problem\citep{choi2025}. Evidence based on this dataset shows that large language models without retrieval perform poorly, while retrieval based approaches improve answer accuracy by grounding responses in relevant content. At the same time, studies using FinDER show that retrieval design is especially important in financial contexts, since standardized language across firms can lead to incorrect retrieval under conventional chunk based approaches\citep{cheng2026a,cheng2026b}. These findings highlight the advantage of RAG for financial question answering tasks that require information beyond a single document.

\section{Problem Identification}

\subsection{Incompleteness of Structured Database}

The incompleteness of structured segment database arises from two distinct sources. First, as shown in Table~\ref{tab:table1}, some firm--year observations that exist in Compustat Fundamental Annual are missing from the Compustat Historical Segment database\footnote{We acknowledge that the absence of segment records for certain firm–years may reflect either database limitations or the reality of a single operating segment. Under ASC Topic 280, firms reporting only one segment are not required to provide disaggregated segment information. Therefore, observed gaps in the Historical Segments database may arise from both structural reporting choices and data coverage limitations.}. Second, even when segment data are available, structured databases often omit nested segment disclosures embedded within the original filings. Together, these two issues create a gap between what firms disclose in their Form 10-K reports and what researchers can observe in standardized datasets.

The first issue concerns missing firm–year observations. For example, Kinder Morgan, Inc. (KMI) appears in Compustat Fundamental Annual for fiscal year 2017 but has no corresponding record in the Historical Segments database. However, upon examining its Form 10-K filing, we find that the firm clearly reports segment-related information\footnote{\url{https://www.sec.gov/Archives/edgar/data/1506307/000150630718000010/kmi-2017x10k.htm\#s52DD36A786795939866455B8530D47AD}}. The absence of this firm–year in the structured segment database therefore does not reflect a lack of disclosure. Missing firm–year observations can create serious concerns for time-series research, as researchers may incorrectly interpret database absence as non-disclosure or structural change, leading to biased inferences about segment evolution over time.

\begin{figure}[p]
\caption{Adobe Inc Segment Disclosure for FY2024}
\label{fig:fig2}
\begin{subfigure}{\textwidth}
\raggedright
Panel A: Reportable Segments
\vspace{0.2cm}
\centering
\includegraphics[width=\textwidth]{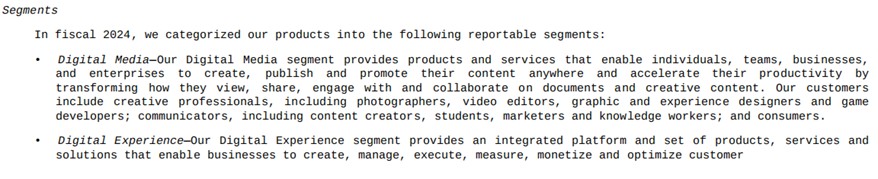}
\vspace{-0.15cm}
\includegraphics[width=\textwidth]{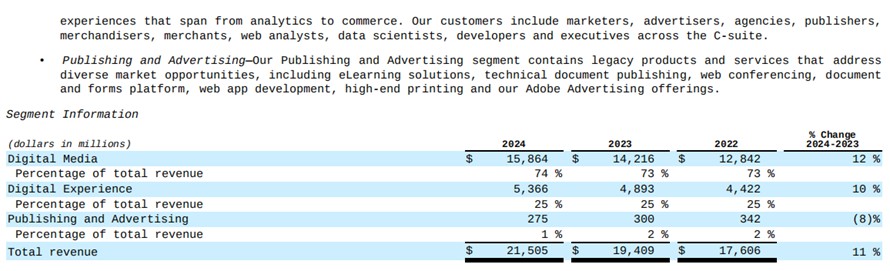}
\end{subfigure}
\vspace{0.6cm}
\begin{subfigure}{\textwidth}
\raggedright
Panel B: Revenue by Major Offerings in Digital Media Reportable Segment
\vspace{0.2cm}
\centering
\includegraphics[width=\textwidth]{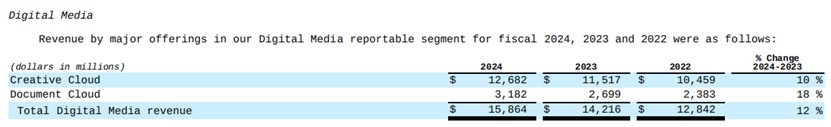}
\end{subfigure}
\end{figure}

Even when firm–year observations are present, structured databases often retain only primary segment identifiers while ignoring nested segment disclosures. Adobe Inc. (Ticker: ADBE) provides a clear illustration. In its fiscal year 2024 Form 10-K, Adobe reports three primary business segments and additionally discloses geographic revenue information. Beyond these dimensions, however, Adobe further disaggregates the Digital Media segment into product-level components such as Creative Cloud and Document Cloud, as shown in Figure 2\ref{fig:fig2}. While business segments are captured in structured databases, these product-level nested disclosures are typically omitted. As a result, researchers relying solely on structured data may overlook economically meaningful lower-tier information embedded within higher-level segment categories.

Together, missing firm–year coverage and omission of nested disclosures show two distinct dimensions of database incompleteness. The first affects sample coverage across time, while the second affects the structural depth of observed disclosures. Both issues motivate returning to primary filings to preserve not only the completeness of firm–year observations but also the hierarchical structure of segment reporting.

\subsection{Comparability Challenges in Segment Disclosure}

A second major challenge in segment disclosure is comparability. Prior literature highlights that segment reporting under the management approach improves relevance but reduces comparability across firms and overtime\citep{hinson2019b}. Unlike the incompleteness issue discussed in Section 3.1, this problem arises even when segment information is available in structured databases, because the way segments are defined, labeled, and reported is inherently discretionary. This creates difficulties for both longitudinal analysis within a firm and comparative analysis across firms.

Prior studies have examined segment changes in relation to economic determinants such as corporate diversification, refocusing, and managerial incentives, including CEO turnover and firm performance\citep{cho2024}. A central issue in this literature is how to identify segment changes over time. Most studies rely on Compustat historical segment identifiers to detect changes in segment structure. While this method provides a convenient operational measure, it does not capture the underlying economic meaning of the change. The same pattern in identifiers may correspond to fundamentally different events, such as mergers and acquisitions, divestitures, internal reorganizations, or simple renaming of existing segments.\citet{denis1997,hyland1997} document that a substantial portion of changes in the number of reported segments reflect reporting adjustments rather than real changes in firm scope. As a result, SID based measures conflate heterogeneous economic events and may lead to misclassification of segment changes.

In addition, SID based approaches do not provide information on how segments relate across periods. When firms reorganize their segment structure, prior segments may be merged, split, or partially reassigned into new segments. However, these relationships are not explicitly recorded in structured databases, which only reflect discontinuities in identifiers. Without understanding how current segments map to prior segments, it is difficult to trace the evolution of a firm’s business structure or to interpret whether observed changes represent substantive restructuring or reporting variation. This limitation makes it challenging to answer longitudinal questions that require integrating information across multiple years of disclosure.

Comparability challenges are further amplified in cross firm settings due to the management approach to segment reporting. Under ASC Topic 280, firms have substantial discretion in defining operating segments based on internal reporting practices, and there is no uniform rule governing how firms aggregate or disaggregate activities. For example, in geographic segment disclosure, one firm may report detailed country level segments such as China, Japan, and Singapore, while another aggregates these activities into a single Asia segment or uses partially overlapping regional definitions. Prior research shows that such discretion can reflect proprietary and agency considerations, including incentives to obscure unfavorable performance or avoid revealing competitively sensitive information\citep{bens2011,berger2007}. These differences are not merely cosmetic, as aggregation decisions affect the visibility and interpretation of performance across segments.
Taken together, these issues show that comparability in segment disclosure is not simply a data availability problem but a structural limitation of how segment information is reported. Addressing this challenge requires approaches that go beyond structured identifiers and incorporate the contextual information contained in firms’ disclosures, enabling more accurate interpretation of segment changes and more consistent alignment of segment structures.

\section{Artifact Development}

\subsection{Data and Sample}

The artifact operates on firms’ annual Form 10-K filings obtained from the SEC’s EDGAR system, which serve as the authoritative source of segment disclosure. Unlike structured databases, the artifact does not rely on pre-existing identifiers, templates, or keyword-based filtering. Instead, it processes complete filings to reconstruct segment information directly from source documents.

To evaluate the artifact, we apply the workflow to 7,653 Form 10-K filings from S\&P 500 firms over the period 2000 to 2025. This setting provides a comprehensive and economically meaningful sample with substantial variation in segment reporting practices, allowing us to assess both extraction completeness and longitudinal consistency.

\subsection{File-Grounded Extraction of Segment Disclosures}
The first component of the artifact addresses the incompleteness of structured databases discussed in Section 3.1. Structured datasets may omit firm-year observations or fail to capture nested segment disclosures embedded in narrative footnotes. To overcome this limitation, the artifact performs direct extraction from primary filings using a file-grounded large language model (LLM) pipeline.

\begin{figure}[p]
\centering
\caption{LLM Assisted Segment Disclosure Extraction Workflow}
\label{fig:fig3}
\includegraphics[width=\textwidth,height=0.82\textheight,keepaspectratio]{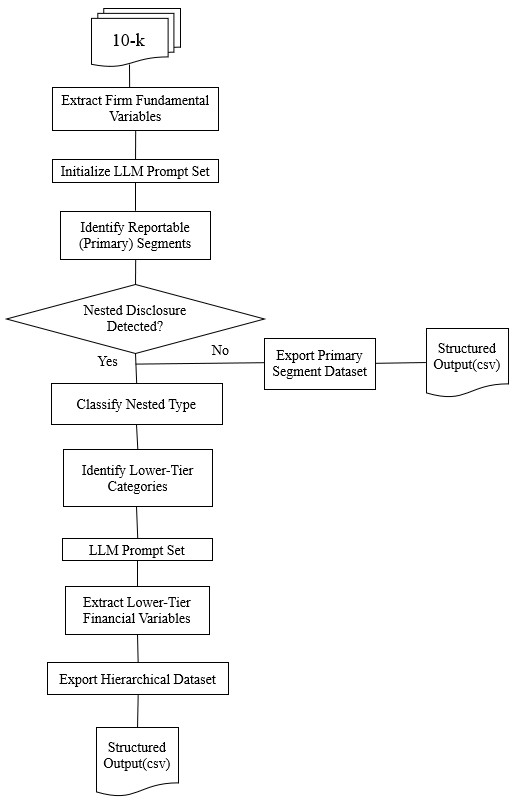}
\end{figure}

Figure 3\ref{fig:fig3} presents the extraction workflow designed to recover segment disclosures directly from Form 10-K filings. For each firm-year, the complete filing obtained from EDGAR serves as the sole authoritative input. The pipeline utilizes the OpenAI API, specifically the Responses API in conjunction with the Files API. Each Form 10-K filing is uploaded once and referenced via a file identifier, allowing multiple extraction queries to be executed efficiently without repeatedly transmitting the full document. The default extraction model is based on the GPT-4.1, which provides strong performance in long-document understanding and structured information extraction. The system further supports concurrent query execution using a thread-based architecture, enabling scalable processing across large samples of filings.

The extraction process proceeds in three stages. First, the model determines whether the firm reports multiple operating segments or operates as a single reporting unit. This classification step ensures that segment extraction is only performed when relevant disclosures are present. Second, for firms with reportable segments, the model extracts segment names and associated financial information by integrating information across both tabular disclosures and narrative footnotes. Third, the workflow evaluates whether nested disclosures are present. Nested disclosures arise when firms provide additional disaggregation within a primary segment, such as product-level, geographic, or revenue-type breakdowns. When such structures are detected, the system triggers an additional extraction stage that captures lower-tier categories and explicitly links them to their parent segments. This branching logic preserves the hierarchical nature of segment reporting, allowing the final output to retain both primary and nested structures rather than collapsing them into a flat dataset.

To ensure consistency and usability of extracted outputs, the system employs a structured prompting design that enforces strict output formatting rules. In particular, the model is instructed to return only the extracted values (rather than full sentences) and to standardize list outputs using delimiters. This design choice reduces post-processing requirements and improves the reliability of downstream analysis. Appendix A\ref{app:appa} presents the prompt template used for file-grounded segment extraction, along with a representative example of the corresponding structured output. Appendix B\ref{app:appb} provides the prompt template and an illustrative example for identifying and extracting nested segment disclosures within reportable segments.

\subsection{Retrieval-Augmented System for Longitudinal and Cross-Firm Comparability}

\begin{figure}[p]
\centering
\caption{RAG System to Improve Comparability of Segment Disclosure}
\label{fig:fig4}
\includegraphics[width=\textwidth]{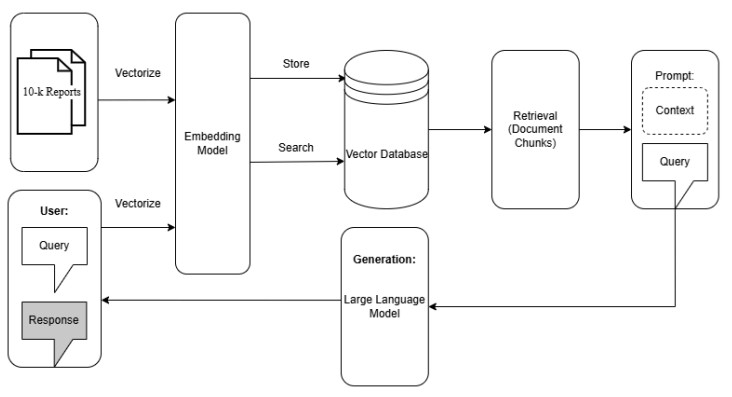}

\begin{minipage}{\textwidth}
\footnotesize \textit{Note:} Adapted from \citet{iaroshev2024}, ``Evaluating Retrieval-Augmented Generation Models for Financial Report Question and Answering,'' \textit{Applied Sciences}, 14(20), 9318.
\end{minipage}
\end{figure}

Figure 4\ref{fig:fig4} illustrates the retrieval-augmented component designed to improve the comparability of segment disclosures across both time and firms. While the extraction workflow recovers segment information for each firm-year, this component focuses on supporting questions that require integrating information across multiple disclosures. 
A key challenge identified in Section 3.2 is that segment disclosures are inherently contextual. Within a firm, segment definitions evolve over time, and changes cannot be interpreted accurately using information from a single year. Across firms, differences in aggregation and labeling make it difficult to directly compare segment structures, even when firms operate in similar economic settings. These challenges motivate the use of a retrieval-augmented design that incorporates information beyond a single document.
To illustrate how the retrieval-augmented system addresses these challenges, we focus on two representative question types. The first is a longitudinal question within a firm, which asks how segment structures evolve over time. In this setting, the system retrieves segment disclosures from multiple years and uses this contextual information to interpret changes in segment definitions, identify whether changes reflect substantive restructuring or reporting variation, and establish relationships between segments across periods.
The second is a cross-firm comparability question, which asks how segment disclosures can be aligned across firms with different reporting structures. In this setting, the system retrieves disclosures from multiple firms and evaluates how segments correspond to each other despite differences in aggregation and naming conventions. Rather than relying on exact matching of labels, the system compares the underlying economic content of disclosures and reconstructs comparable segment groupings.

\section{Evaluation}

\subsection{Evaluation of File-Grounded Segment Extraction}
To evaluate the effectiveness of the proposed workflow, we conduct a manual verification using S\&P 500 firms’ 10-K filings from fiscal years 2000 to 2025. The full corpus contains 7,653 filings. For each evaluation round, we randomly sample 30 filings and repeat the process three times to ensure robustness. The results are in Table 2\ref{tab:table2}.

We first manually examine the 30 sampled filings in each round to determine whether firms operate as a single reporting unit or disclose multiple reportable segments. The verification results show that 27, 25, and 22 firms in Groups 1 to 3, respectively, disclose multiple reportable segments rather than operating as a single reporting unit. We then compare these manual labels with the identification results produced by ChatGPT. The model correctly identifies 27, 25, and 22 filings with multi-segment disclosure across the three groups, indicating perfect classification consistency with manual verification in detecting the presence of reportable segments.

Next, we evaluate the accuracy of financial information extraction for reportable (primary) segments. For each evaluation round, we randomly select 100 extracted cells corresponding to segment-level financial information and manually compare them with values reported in the original filings. The average extraction accuracy rates for reportable segments are 97 percent, 91 percent, and 94 percent for Groups 1 to 3, respectively, demonstrating strong performance in extracting primary segment financial data.

We then examine nested segment disclosure. Nested disclosure refers to situations where firms provide additional breakdown under a reportable segment. For example, as illustrated in Figure 2\ref{fig:fig2}, Adobe Inc. in fiscal year 2024 reports three reportable segments, Digital Media, Digital Experience, and Publishing and Advertising, and further discloses product-level information within the Digital Media segment, which constitutes nested segment disclosure. For the same 30 filings in each group, we manually determine whether firms provide additional disaggregation within reportable segments. The verification shows that 13, 13, and 15 filings contain nested segment disclosure across Groups 1 to 3. ChatGPT identifies 14, 16, and 15 filings with nested disclosure, indicating slightly higher detection counts in two groups but overall close alignment with manual verification. 

Finally, we evaluate the extraction accuracy of nested segment financial information. Similar to the primary segment evaluation, we randomly select 100 extracted cells associated with nested disclosures and manually verify their correctness. The average extraction accuracy rates for nested segment financial information are 86 percent, 77 percent, and 88 percent across Groups 1 to 3.

Overall, the evaluation results indicate that the proposed workflow reliably identifies multi-segment disclosure and achieves high accuracy in extracting financial information for both reportable and nested segments across repeated random samples.

\subsection{Evaluation of Retrieval-Augmented Segment Change Identification}
We illustrate the capability of the retrieval-augmented component using two representative questions that reflect the comparability challenges discussed in Section 3.2: a longitudinal question within a firm and a cross-firm comparability question. These examples are intended to demonstrate how the system leverages information across multiple disclosures to support interpretation, rather than to provide a comprehensive evaluation of performance.

The first example considers a longitudinal question that asks how a firm’s segment structure evolves over time. Table 3\ref{tab:table3} presents the output for Avery Dennison (AVY)\footnote{Avery Dennison is selected as an illustrative case because it exhibits multiple types of segment changes over the sample period, including internal reorganizations, divestitures, and reporting-driven reclassification. These features provide a rich setting to demonstrate how the retrieval-augmented system interprets segment evolution and establishes linkages across periods.}, where the system processes Form 10-K filings over the period 2000–2024 and constructs a structured summary of segment changes. For each year, the table reports reportable segment names, an indicator of whether a change occurred, the inferred reason for the change, and the linkage between current and prior segments.

This example illustrates how retrieval across multiple years enables the system to interpret segment changes in context. Rather than relying on mechanical differences in segment identifiers, the system incorporates information from prior disclosures to distinguish between continuity and change and to provide economically meaningful interpretations, such as internal reorganization, divestiture, or reporting-driven reclassification. In addition, the system identifies how segments relate across periods, capturing relationships such as continuation, merging, or splitting, which are not directly observable from structured identifiers.

The second example considers a cross-firm comparability question that asks how geographic segment disclosures can be aligned across firms. Table 4\ref{tab:table4}  presents the results for Intel (INTC) and Texas Instruments (TXN)\footnote{Intel and Texas Instruments are selected because both firms disclose substantial geographic segment information in Asia but adopt different aggregation strategies. Intel reports relatively disaggregated country-level segments, while Texas Instruments uses more aggregated regional categories and modifies its reporting structure over time. In addition, both firms operate in the same industry based on their Standard Industrial Classification (SIC) codes, and are therefore more likely to be grouped together in empirical research settings. This provides a relevant context to illustrate how the retrieval-augmented system aligns segment disclosures across firms with different reporting structures.}, where the system reconstructs comparable geographic exposure in Asia despite differences in reporting structures. Intel reports disaggregated country-level segments, while Texas Instruments reports more aggregated regional categories.

This example demonstrates how the retrieval-augmented system reconciles differences in aggregation by mapping firm-specific disclosures into comparable groupings. The system aggregates detailed country-level disclosures for Intel into a unified regional measure and aligns them with the broader categories reported by Texas Instruments, while preserving the underlying segment composition and associated financial information. This allows consistent cross-firm comparison without discarding firm-specific detail.

Taken together, these two examples illustrate how retrieval-augmented generation can support both longitudinal interpretation and cross-firm alignment of segment disclosures. By incorporating contextual information from multiple documents, the approach provides a flexible framework for addressing comparability challenges that are difficult to resolve using structured data alone.

\section{Conclusion}

Segment-level disclosures play a critical role in financial reporting by providing insights into firms’ internal organization and the allocation of economic activities across operating units. Segment disclosures are reported in firms’ SEC filings, particularly in Form 10-K filings, where they appear in both narrative and tabular formats and reflect managerial discretion under the management approach. These features limit the ability of structured databases to fully capture segment information and create challenges for both completeness and comparability in empirical research.

This study shows that large language models can serve as a useful tool for extracting segment disclosures directly from Form 10-K filings, including both reportable segments and nested information embedded within disclosures. In addition, we demonstrate that a retrieval-augmented design can support comparability by incorporating contextual information across multiple documents. Using two illustrative questions, we show how the approach enables interpretation of segment changes within a firm over time and alignment of segment disclosures across firms despite differences in reporting structures.

This study has several limitations that suggest directions for future research. First, the current analysis focuses on S\&P 500 firms, which provide a large and economically meaningful sample but may limit generalizability. Future research may extend the approach to a broader set of publicly listed firms, including smaller firms and firms with less standardized disclosure practices, to evaluate the scalability and robustness of the proposed framework. Second, this study employs the GPT-4.1-mini model, future work may explore the use of alternative models or architectures to further improve extraction accuracy and interpretability. Third, while this study focuses on two representative comparability questions, the framework can be extended to a wider range of research settings. Future work may explore additional questions that require cross-period or cross-firm contextual understanding, such as tracking product-level evolution within segments, analyzing changes in geographic exposure under alternative aggregation schemes, or examining how segment reporting responds to regulatory or strategic shifts.

More broadly, this study highlights the potential of large language models as a complementary tool for extracting and interpreting information from unstructured financial disclosures. By returning to primary filings and incorporating contextual information across documents, the proposed approach provides a flexible foundation for improving the measurement and comparability of segment-level data in accounting research.

\newpage

\bibliographystyle{apalike}
\bibliography{references}

\newpage

\begin{longtable}{p{2cm} p{12cm}}
\caption{ Missing Firm-year Segment Data From Compustat for S\&P 500 Firms}
\label{tab:table1} \\

\toprule
\textbf{Year} & \textbf{CIK} \\
\midrule
\endfirsthead

\toprule
\textbf{Year} & \textbf{CIK} \\
\midrule
\endhead

1993 & 1390777, 19617, 28412, 35527, 1281761, 36270, 36104, 49196, 70858, 73124, 72971, 713676, 91576, 93751, 109380, 92230, 1132979, 1378946, 719739 \\

1994 & 1390777, 19617, 28412, 35527, 1281761, 36270, 36104, 49196, 70858, 73124, 72971, 713676, 91576, 93751, 109380, 92230, 1132979, 1378946, 719739 \\

1995 & 1390777, 19617, 28412, 35527, 1281761, 36270, 36104, 49196, 70858, 73124, 72971, 713676, 91576, 93751, 109380, 92230, 1132979, 1378946, 719739 \\

1996 & 1390777, 19617, 28412, 35527, 1281761, 36270, 36104, 49196, 70858, 73124, 72971, 713676, 91576, 93751, 109380, 92230, 1132979, 1378946, 719739, 1521332 \\

1997 & 1390777, 19617, 28412, 36270, 35527, 1281761, 36104, 49196, 70858, 73124, 72971, 713676, 91576, 93751, 109380, 92230, 1132979, 1378946, 719739, 1521332 \\

1998 & 1390777, 19617, 28412, 36270, 35527, 1281761, 36104, 49196, 70858, 73124, 72971, 713676, 91576, 93751, 109380, 92230, 1132979, 1378946, 719739, 1001039 \\

1999 & 1390777, 19617, 28412, 36270, 35527, 1281761, 36104, 49196, 70858, 73124, 72971, 713676, 91576, 93751, 109380, 92230, 1132979, 1378946, 719739 \\

2000 & 1390777, 19617, 28412, 36270, 35527, 1281761, 36104, 49196, 70858, 73124, 72971, 713676, 91576, 93751, 109380, 92230, 1132979, 1378946, 719739, 1156375 \\

2001 & 1390777, 19617, 28412, 36270, 35527, 1281761, 36104, 49196, 70858, 73124, 72971, 713676, 91576, 93751, 109380, 92230, 1132979, 1378946, 719739 \\

2002 & 1390777, 19617, 28412, 36270, 35527, 1281761, 36104, 49196, 70858, 73124, 72971, 713676, 91576, 93751, 109380, 92230, 1132979, 1378946, 719739 \\

2003 & 1390777, 19617, 35527, 36270, 1281761, 36104, 49196, 70858, 73124, 72971, 91576, 93751, 109380, 92230, 1132979, 1378946, 719739 \\

2007 & 1585689, 1506307, 1402057 \\
2008 & 1585689, 1506307 \\
2009 & 1585689, 1506307 \\
2010 & 1585689, 1506307 \\
2011 & 1506307 \\
2012 & 1506307 \\
2013 & 1506307, 1699150 \\
2014 & 1506307, 1699150 \\
2015 & 1506307, 1701605, 1751788, 1755672, 1781335 \\
2016 & 1506307, 1701605, 1755672, 1781335 \\
2017 & 1506307, 1781335 \\
2018 & 1506307 \\
2019 & 1506307 \\
2020 & 1506307 \\
2021 & 1506307 \\
2022 & 1506307 \\
2023 & 1506307 \\
2024 & 1506307 \\

\bottomrule
\multicolumn{2}{p{14cm}}{\footnotesize \textit{Note:} We construct this table by matching firm-year observations across Compustat Fundamental Annual (FUNDA) and Compustat Historical Segment data to identify years in which firms have accounting information but lack corresponding segment-level disclosures. Our analysis begins in fiscal year 1993, the earliest year in which Form 10-K filings are broadly available through the SEC’s EDGAR system. Specifically, we merge firm identifiers using CIKs and fiscal years, and flag firm-year observations present in Fundamental Annual but missing from the Segment History files. Aggregating these missing observations by calendar year yields 17,993 firm-year observations for all U.S.-listed firms. When restricting the sample to S\&P 500 constituent firms, the number of missing firm-year observations decreases to 266. The table reports, for each year, the set of CIKs associated with these missing segment disclosure} \\
\end{longtable}

\begin{table}[htbp]
\centering
\caption{Evaluation of Extraction Results}
\label{tab:table2}

\begin{tabular}{lccc}
\toprule
 & \textbf{Group 1} & \textbf{Group 2} & \textbf{Group 3} \\
\midrule

Num of 10-K Filings & 30 & 30 & 30 \\

Num of Firms with Multi-Segment Disclosure & 27 & 25 & 22 \\

ChatGPT Identified Multi-Segment Filings & 27 & 25 & 22 \\

Primary Segment Extraction Accuracy (\%) & 97\% & 91\% & 94\% \\

Num of Observations with Nested Disclosure & 13 & 13 & 15 \\

ChatGPT Identified Nested Disclosure & 14 & 16 & 15 \\

Nested Segment Extraction Accuracy (\%) & 86\% & 77\% & 88\% \\

\bottomrule
\end{tabular}

\end{table}

\begin{landscape}

\small
\begin{longtable}{p{1.5cm} p{4.5cm} p{1.5cm} p{5cm} p{4cm}}

\caption{RAG-based Segment Change Identification for Avery Dennison Corp}
\label{tab:table3} \\

\toprule
\textbf{Year} & \textbf{Reportable Segment Name(s)} & \textbf{Change?} & \textbf{Reason for Change} & \textbf{Linked with Prior Segment?} \\
\midrule
\endfirsthead

\toprule
\textbf{Year} & \textbf{Reportable Segment Name(s)} & \textbf{Change?} & \textbf{Reason for Change} & \textbf{Linked with Prior Segment?} \\
\midrule
\endhead

2001 & Pressure-sensitive Adhesives and Materials; Consumer and Converted Products & No &  &  \\

2002 & Pressure-sensitive Adhesives and Materials; Consumer and Converted Products & No &  &  \\

2003 & Pressure-sensitive Adhesives and Materials; Consumer and Converted Products & No &  &  \\

2004 & Pressure-sensitive Materials; Office Products; Other Converted Products and Services; Retail Information Services & Yes & Internal reorganization / reporting-driven change (growth of retail information services >10\%) & Partial (Consumer \& Converted Products split; PSM continues) \\

2005 & Pressure-sensitive Materials; Office and Consumer Products; Retail Information Services & Yes & Internal reclassification (consolidation from 4 to 3 segments) & Yes (re-grouping; RIS and PSM continue) \\

2006 & Pressure-sensitive Materials; Office and Consumer Products; Retail Information Services & No &  &  \\

2007 & Pressure-sensitive Materials; Office and Consumer Products; Retail Information Services & No &  &  \\

2008 & Pressure-sensitive Materials; Retail Information Services; Office and Consumer Products & No &  &  \\

2009 & Pressure-sensitive Materials; Retail Information Services; Office and Consumer Products & No &  &  \\

2010 & Pressure-sensitive Materials; Retail Information Services; Office and Consumer Products & No &  &  \\

2011 & Pressure-sensitive Materials; Retail Information Services; Office and Consumer Products & No &  &  \\

2012 & Pressure-sensitive Materials; Retail Branding and Information Solutions & Yes & Internal reorganization + divestiture (OCP removed) & Partial (PSM continues; RIS to RBIS; OCP discontinued) \\

2013 & Pressure-sensitive Materials; Retail Branding and Information Solutions & No &  &  \\

2014 & Pressure-sensitive Materials; Retail Branding and Information Solutions; Vancive Medical Technologies & Yes & Vancive became a separate reportable segment & Partial (PSM and RBIS continue; Vancive added) \\

2015 & Pressure-sensitive Materials; Retail Branding and Information Solutions; Vancive Medical Technologies & No &  &  \\

2016 & Label and Graphic Materials; Retail Branding and Information Solutions; Industrial and Healthcare Materials & Yes & Internal reorganization; segment components reallocated & Partial (RBIS continues; PSM reorganized; Vancive absorbed) \\

2017 & Label and Graphic Materials; Retail Branding and Information Solutions; Industrial and Healthcare Materials & No &  &  \\

2018 & Label and Graphic Materials; Retail Branding and Information Solutions; Industrial and Healthcare Materials & No &  &  \\

2019 & Label and Graphic Materials; Retail Branding and Information Solutions; Industrial and Healthcare Materials & No &  &  \\

2020 & Label and Graphic Materials; Retail Branding and Information Solutions; Industrial and Healthcare Materials & No &  &  \\

2021 & Label and Graphic Materials; Retail Branding and Information Solutions; Industrial and Healthcare Materials & No &  &  \\

2022 & Materials Group; Solutions Group & Yes & Internal reorganization; segments consolidated and renamed & Partial (Materials Group from LGM + IHM; Solutions from RBIS) \\

2023 & Materials Group; Solutions Group & No &  &  \\

2024 & Materials Group; Solutions Group & No &  &  \\

\bottomrule
\end{longtable}
\normalsize
\end{landscape}
\begin{landscape}
\scriptsize
\begin{longtable}{p{1cm} p{3cm} p{2cm} p{3cm} p{3cm} p{2cm} p{2cm} p{2cm} p{2cm}}
\caption{RAG-based Alignment of Geographic Segments Across Firms}
\label{tab:table4} \\
\toprule
\textbf{Year} & \textbf{Segments in Asia for INTC} & \textbf{Segments in Asia for TXN} & \textbf{Detailed Segment Performance for INTC} & \textbf{Detailed Segment Performance for TXN} & \textbf{Sales for INTC in Asia} & \textbf{Sales for TXN in Asia} & \textbf{\% Asia / Total INTC} & \textbf{\% Asia / Total TXN} \\
\midrule
\endfirsthead

\toprule
\textbf{Year} & \textbf{Segments in Asia for INTC} & \textbf{Segments in Asia for TXN} & \textbf{Detailed Segment Performance for INTC} & \textbf{Detailed Segment Performance for TXN} & \textbf{Sales for INTC in Asia} & \textbf{Sales for TXN in Asia} & \textbf{\% Asia / Total INTC} & \textbf{\% Asia / Total TXN} \\
\midrule
\endhead

2012 & Singapore, China incl.\ HK, Taiwan, Japan & Asia, Japan & Singapore, 12,622; China incl.\ HK, 8,299; Taiwan, 9,327; Japan, 4,303 & Asia, 7,808; Japan, 1,357 & 34,551 & 9,165 & 64.8\% & 71.5\% \\

2013 & Singapore, China incl.\ HK, Taiwan, Japan & Asia, Japan & Singapore, 10,997; China incl.\ HK, 9,890; Taiwan, 8,888; Japan, 3,725 & Asia, 7,370; Japan, 1,072 & 33,500 & 8,442 & 63.6\% & 69.2\% \\

2014 & Singapore, China incl.\ HK, Taiwan, Japan & Asia, Japan & Singapore, 11,573; China incl.\ HK, 11,197; Taiwan, 8,955; Japan, 2,776 & Asia, 7,915; Japan, 1,032 & 34,501 & 8,947 & 61.8\% & 68.6\% \\

2015 & Singapore, China incl.\ HK, Taiwan & Asia, Japan & Singapore, 11,544; China incl.\ HK, 11,679; Taiwan, 10,350 & Asia, 7,910; Japan, 1,127 & 33,573 & 9,037 & 60.7\% & 69.5\% \\

2016 & Singapore, China incl.\ HK, Taiwan & Asia, Japan & Singapore, 12,780; China incl.\ HK, 13,977; Taiwan, 9,953 & Asia, 8,024; Japan, 1,040 & 36,710 & 9,064 & 61.8\% & 67.8\% \\

2017 & Singapore, China incl.\ HK, Taiwan & Asia, Japan & Singapore, 14,285; China incl.\ HK, 14,796; Taiwan, 10,518 & Asia, 8,824; Japan, 1,049 & 39,599 & 9,873 & 63.1\% & 66.0\% \\

2018 & Singapore, China incl.\ HK, Taiwan & Asia, Japan & Singapore, 15,409; China incl.\ HK, 18,824; Taiwan, 10,646 & Asia, 9,240; Japan, 869 & 44,879 & 10,109 & 63.3\% & 64.1\% \\

2019 & Singapore, China incl.\ HK, Taiwan & Asia, Japan & Singapore, 15,650; China incl.\ HK, 20,026; Taiwan, 10,058 & Asia, 8,650; Japan, 796 & 45,734 & 9,446 & 63.6\% & 65.7\% \\

2020 & Singapore, China incl.\ HK, Taiwan & Asia, Japan & Singapore, 17,845; China incl.\ HK, 20,257; Taiwan, 11,605 & Asia, 9,541; Japan, 734 & 49,707 & 10,275 & 63.8\% & 71.1\% \\

2021 & Singapore, China incl.\ HK, Taiwan & Asia, Japan & Singapore, 18,096; China incl.\ HK, 22,961; Taiwan, 11,418 & China, 4,586; Rest of Asia, 2,018; Japan, 1,468 & 52,475 & 8,072 & 66.4\% & 71.7\% \\

2022 & Singapore, China incl.\ HK, Taiwan & China, Rest of Asia, Japan & Singapore, 9,664; China incl.\ HK, 17,125; Taiwan, 8,287 & China, 4,807; Rest of Asia, 2,003; Japan, 1,602 & 35,076 & 8,412 & 55.6\% & 42.0\% \\

2023 & Singapore, China incl.\ HK, Taiwan & China, Rest of Asia, Japan & Singapore, 8,602; China incl.\ HK, 14,854; Taiwan, 6,867 & China, 3,293; Rest of Asia, 1,721; Japan, 1,782 & 30,323 & 6,796 & 55.9\% & 38.8\% \\

2024 & Singapore, China incl.\ HK, Taiwan & China, Rest of Asia, Japan & Singapore, 10,187; China incl.\ HK, 15,532; Taiwan, 7,804 & China, 3,012; Rest of Asia, 1,681; Japan, 1,212 & 33,523 & 5,905 & 63.1\% & 37.8\% \\

\bottomrule
\end{longtable}
\normalsize

\end{landscape}

\clearpage
\appendix
\section*{Appendix A: Example of Prompts and Extraction Results}
\label{app:appa}
\addcontentsline{toc}{section}{Appendix A: Example of Prompts and Extraction Results}
Appendix A provides illustrative examples of the prompts and responses\footnote{As documented by\citep{cai2026}, neutral or highly polite prompts generally yield higher accuracy in modern large language models. Accordingly, we use a neutral tone in our prompt design to help promote consistent and reliable responses.} used in the extraction workflow described in Section 4.2. The appendix demonstrates how the artifact interacts with the source Form 10-K filing to retrieve structured information through guided prompts. For each variable category, the table reports the variable name, the prompt presented to the model, and the corresponding output generated by the system. These examples illustrate how the artifact translates unstructured disclosure content into standardized, machine-readable fields that can be used for subsequent analysis.
\small
\begin{longtable}{p{0.7cm}p{2cm}p{1.8cm}p{7cm}p{3cm}}
\toprule
\textbf{\#} & \textbf{Category} & \textbf{Variable} & \textbf{Prompt} & \textbf{Result}\\
\midrule
\endfirsthead
\toprule
\textbf{\#} & \textbf{Category} & \textbf{Variable} & \textbf{Prompt} & \textbf{Result}\\
\midrule
\endhead
1 & general & gvkey & What is the GVKEY for the firm in this year? & 6614\\
2 & general & conm & What is the exact legal name of the firm for this year? & Apple Inc.\\
3 & general & tic & What stock ticker is associated with the firm for this year? & AAPL\\
4 & general & cik & What SEC CIK number is associated with the firm in this year? & 320193\\
5 & general & sic & What is the primary SIC code for the geographic segment of the firm in this year? & 3571\\
6 & general & sics1 & Does the geographic segment have a different SIC or industry code from the consolidated firm in this year? & No\\
7 & general & sics2 & What additional SIC classification is reported for the geographic segment in this year? & 7372\\
8 & general & naics & What is the primary NAICS code for the firm in this year? & 334111\\
9 & general & naicsh & What NAICS hierarchy or description is provided for the firm in this year? & Not provided\\
10 & general & naicss1 & What NAICS code is associated with the firm in this year? & 334111\\
11 & general & naicss2 & What additional NAICS classification is reported for the firm in this year? & 334220\\
12 & general & gind & What GICS industry does the firm belong to in this year? & Technology Hardware, Storage \& Peripherals\\
13 & general & gsubind & What GICS sub-industry does the firm belong to in this year? & Technology Hardware, Storage \& Peripherals\\
14 & general & curcds & In what currency are the segment amounts for the firm in this year presented? & U.S. dollars\\
15 & general & isosrc & What is the ISO currency source or reference associated with the segment disclosures for this year? & U.S. dollar\\
16 & general & srcs & What is the source document used for the firm? & Form 10-K\\
17 & general & revt & What is the consolidated total revenue of the firm in this year? & \$391,035 million\\
\bottomrule
\end{longtable}
\normalsize

\clearpage
\begin{landscape}
\section*{Appendix B: Example for Firm with Nested Disclosure}
\label{app:appb}
\addcontentsline{toc}{section}{Appendix B: Example for Firm with Nested Disclosure}
Appendix B provides an example of the extraction output for a firm that contains nested segment disclosures. The example illustrates how the artifact identifies reportable segments and subsequently detects additional layers of disaggregation within those segments. By linking nested disclosures to their corresponding reportable segment, the workflow preserves the hierarchical structure of segment reporting that often appears in Form 10-K filings. This example demonstrates how the artifact captures both primary and lower-tier segment information from the original filing and organizes the results into a structured representation suitable for empirical analysis.
\vspace{0.4cm}
\begin{center}
\includegraphics[width=1.5\textwidth,height=1.80\textheight,keepaspectratio]{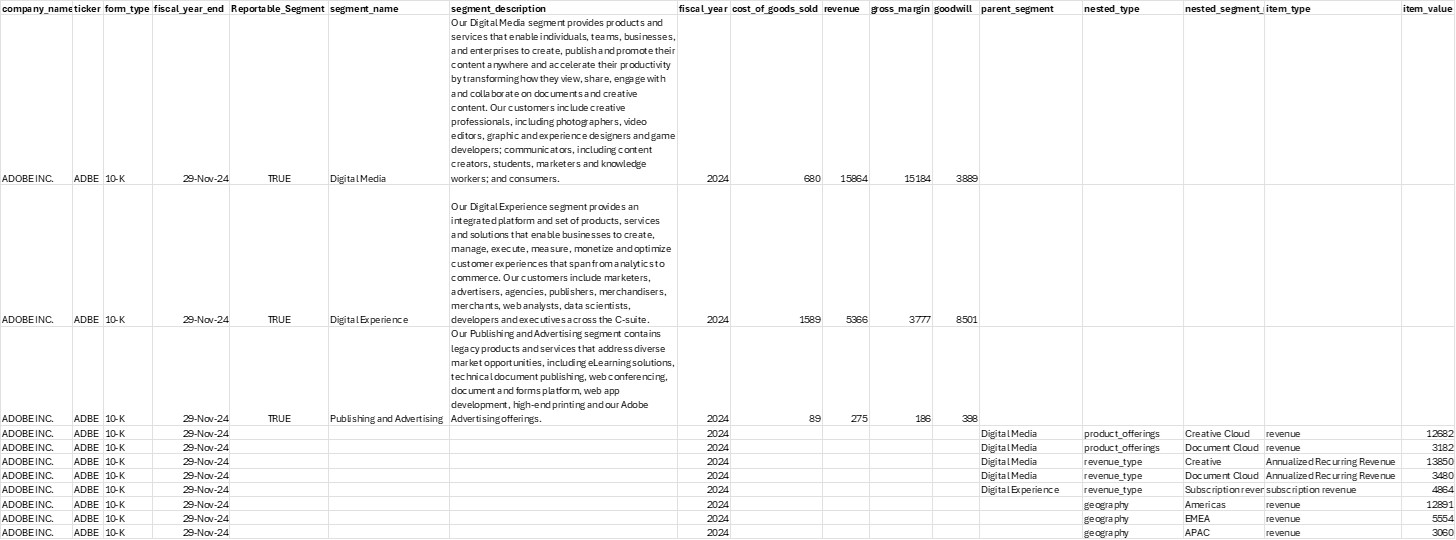}
\end{center}
\end{landscape}

\end{document}